\documentclass[conference]{IEEEtran}
\IEEEoverridecommandlockouts
\usepackage{cite}
\usepackage{amsmath,amssymb,amsfonts}
\usepackage[paper=letterpaper,margin=0.75in]{geometry}
\usepackage{algorithmic}
\usepackage{graphicx}
\usepackage{textcomp}
\usepackage{xcolor}
\usepackage{booktabs, textcomp}
\usepackage{subfigure, caption}

\def\BibTeX{{\rm B\kern-.05em{\sc i\kern-.025em b}\kern-.08em
    T\kern-.1667em\lower.7ex\hbox{E}\kern-.125emX}}

\newcommand{\dmidrule}{\specialrule{0.5pt}{0pt}{0.4pt}%
	\specialrule{0.2pt}{0pt}{\belowrulesep}%
}

\usepackage{float}
\usepackage{aliascnt}
\newaliascnt{eqfloat}{equation}
\newfloat{eqfloat}{h}{eqflts}
\floatname{eqfloat}{Equation}

\newcommand*{\ORGeqfloat}{}
\let\ORGeqfloat\eqfloat
\def\eqfloat{%
	\let\ORIGINALcaption\caption
	\def\caption{%
		\addtocounter{equation}{-1}%
		\ORIGINALcaption
	}%
	\ORGeqfloat
}

\begin{document}
\newgeometry{top=1in,bottom=0.75in,right=0.75in,left=0.75in}
\title{\LARGE Learning When to Ask for Help: Efficient Interactive Navigation via Implicit Uncertainty Estimation}
\author{\IEEEauthorblockN{Ifueko Igbinedion}
	\IEEEauthorblockA{\textit{Laboratory for Information and Decision Systems} \\
		\textit{Massachusetts Institute of Technology}\\
		Cambridge, MA, USA \\
		ifueko@mit.edu}
	\and
	\IEEEauthorblockN{Sertac Karaman}
	\IEEEauthorblockA{\textit{Laboratory for Information and Decision Systems} \\
		\textit{Massachusetts Institute of Technology}\\
		Cambridge, MA, USA \\
		sertac@mit.edu}
}

\maketitle

\begin{abstract}
Robots operating alongside humans often encounter unfamiliar environments that make autonomous task completion challenging. Though improving models and increasing dataset size can enhance a robot's performance in unseen environments, data collection and model refinement may be impractical in every environment. Approaches that utilize human demonstrations through manual operation can aid in refinement and generalization, but often require significant data collection efforts to generate enough demonstration data to achieve satisfactory task performance. Interactive approaches allow for humans to provide correction to robot action in real time, but intervention policies are often based on explicit factors related to state and task understanding that may be difficult to generalize. Addressing these challenges, we train a lightweight interaction policy that allows robots to decide when to proceed autonomously or request expert assistance at estimated times of uncertainty. An implicit estimate of uncertainty is learned via evaluating the feature extraction capabilities of the robot's visual navigation policy. By incorporating part-time human interaction, robots recover quickly from their mistakes, significantly improving the odds of task completion. Incorporating part-time interaction yields an increase in success of 0.38 with only a 0.3 expert interaction rate within the Habitat simulation environment using a simulated human expert. We further show success transferring this approach to a new domain with a real human expert, improving success from less than 0.1 with an autonomous agent to 0.92 with a 0.23 human interaction rate. This approach provides a practical means for robots to interact and learn from humans in real-world settings.
\end{abstract}

\begin{IEEEkeywords}
human factors and human-in-the-loop, vision-based navigation, reinforcement learning
\end{IEEEkeywords}

\section{Introduction}
Large scale simulation environments have enabled the development of increasingly proficient reinforcement learning policies for robotic control.
Online optimization approaches and the wide variety of high-resolution sensors and datasets allow robots to learn complex tasks, including navigation \cite{wijmans2019dd,zeng2020survey}, manipulation \cite{gu2016deep} and vision-language navigation \cite{anderson2018vision}.
Like many applied systems, policies optimized in online simulation environments often struggle when deployed in the real world, especially in domains unsuitable for simulation \cite{levine2020offline}.

Offline imitation learning approaches are aimed at teaching robots to complete tasks by imitating human behavior.
While these approaches achieve impressive performance on a variety of tasks \cite{argall2009survey,wang2019reinforced,hussein2017imitation,ho2016generative} such datasets can be difficult to generate if full teleoperation is required \cite{clever2021assistive,dass2022pato}.
Interactive imitation learning approaches \cite{kelly2019hg,celemin2022interactive} enable a human expert to intervene autonomous robots in real time, enabling the correction of sub-optimal or risky behaviors that may lead to failure in task completion.
However, these approaches may require the explicit definition of criteria for intervention based on state or task understanding that may be difficult to perform generally \cite{zhang2016query,hoque2021thriftydagger}.

\begin{figure}
	\centering
	\begin{subfigure}
		\centering
		\includegraphics[width=0.42\textwidth]{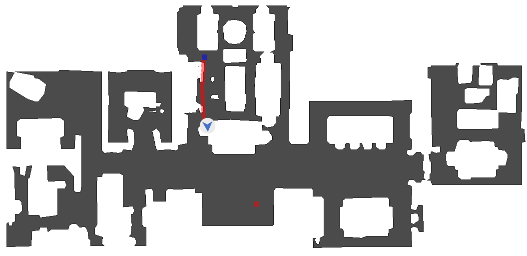}
		\caption*{(a) Autonomous execution without human intervention. In unfamiliar environments, agents may be confused by simple obstacles that result in complete task failure.}
	\end{subfigure}
	\begin{subfigure}
		\centering
		\includegraphics[width=0.42\textwidth]{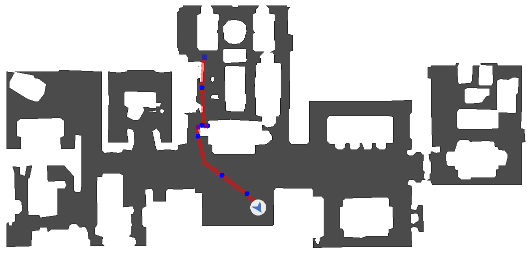}
		\caption*{(b) With uncertainty driven part-time intervention, humans easily prevent task failure.}
	\end{subfigure}
	\caption{The impact of part-time intervention on autonomous agent execution. Points of semi-autonomous control are shown in blue, while autonomous execution is shown in red. When agents are empowered with the ability to decide when to ask for help at points of uncertainty, navigating difficult areas within new environments becomes easy.}
	\label{fig:execution}
\end{figure}
\restoregeometry
In this work, we present an approach to interactive imitation learning in which robot initiated intervention is based on an implicit estimate of uncertainty of the features extracted from the autonomous navigation policy.
Given a pretrained point navigation policy, we train a secondary interaction policy that allows robots choose when to request human input during uncertainty in task execution.
Our most optimal approach yields a 0.38 increase in success weighted path length during interactive control with a 0.3 simulated expert interaction rate in the Habitat simulation environment \cite{habitat19iccv}.
We are further able to transfer policies trained using this approach to a completely unique environment collaborating with a real human expert, showing the zero-shot transfer of our semi-autonomous policy in the FlightGoggles \cite{guerra2019flightgoggles} simulation environment.
While the autonomous agent navigates successfully in this unique environment at a rate of less than 0.1, part-time interaction yields a success rate of 0.92 with just a 0.23 interaction rate.

\section{Related Work}
\subsection{Human-in-the-loop Robotic Planning}
When navigating unfamiliar environments, incorporating humans within the planning loop can prevent catastrophic failure in unexpected scenarios.
Simple approaches allow humans to remotely supervise execution and assume control of robot agents during emergency situations or when changes to autonomous programming are required \cite{sheridan1986human}.

More sophisticated approaches incorporate modeling human or robot behavior for more optimized control, such as \cite{wang2018trust}, which incorporates a model of human trust to optimize the performance of autonomous robots while minimizing operator overload, and \cite{papallas2022ask}, which presents an approach to modeling a robot agent's needs, allowing systems to dispatch a human agent to specific robots in an optimized fashion.
Incorporating just a single part-time operator can allow for significant improvement in task performance for multiple, simultaneously executing robots, suggesting that even minimal human intervention can yield significant performance improvements.

\subsection{Offline Imitation Learning}
Methods that refine control policies from human demonstration have been successful in transferring new behaviors or refining known behaviors in new environments \cite{argall2009survey}.
Direct demonstration techniques for navigation allow humans to provide manual control to teach agents through supervised imitation learning approaches such as behavioral cloning \cite{torabi2018behavioral}.
Once a demonstration dataset is collected from manual human operation, the desired policy can be optimized to minimize the error between predictions and human actions within the demonstration.

Though human demonstrations may streamline control policy optimization, low-quality demonstrations can negatively impact performance of imitation learning policies \cite{hussein2017imitation}. 
To combat this challenge, the approach proposed in \cite{ramrakhya2023pirlnav} leverages the power of imitation learning to pretrain a proficient navigation policy and fine-tunes this policy in reinforcement learning to achieve optimal performance for the task of object navigation.
This approach leverages the availability of both human demonstrations and a live simulation environment to first learn human behaviors that are not easily derived from trial and error before fine tuning those behaviors to achieve optimal policy performance, reducing the need for error-free human demonstrations when simulation environments are available.

\subsection{Interactive Imitation Learning}
While offline imitation learning approaches can fill the gap when simulation environments are unavailable, they also suffer from similar domain adaptation issues as policies trained in the online setting.
In particular, these policies may struggle to generalize to scenarios that deviate too largely from demonstrations in the offline dataset \cite{argall2009survey}.
Interactive imitation learning enables humans to provide actions to autonomous agents in the online setting, allowing for intervention and correction in real time \cite{kelly2019hg,celemin2022interactive,zhang2016query,hoque2021thriftydagger}.
These approaches require an interaction policy that determines at which points in time a human expert should take action.
This policy can be human gated \cite{kelly2019hg} in which the human decides when interaction occurs, or robot gated \cite{celemin2022interactive,zhang2016query,hoque2021thriftydagger} in which the robot initiates the interaction.
Human gated policies can be subpotimal because the human must continuously monitor robot execution and trigger the intervention.
Robot gated policies remove this requirement, but instead must define criteria for intervention.
Some approaches estimate the need for intervention based on task performance \cite{ross2014reinforcement}, state novelty or risk \cite{menda2019ensembledagger,hoque2021thriftydagger}, or state-based uncertainty \cite{menda2019ensembledagger}.
We instead learn an implicit model of uncertainty based on the robot's autonomous policy, estimated by its feature extraction power.

\begin{figure}
	\centering
	\includegraphics[width=0.49\textwidth]{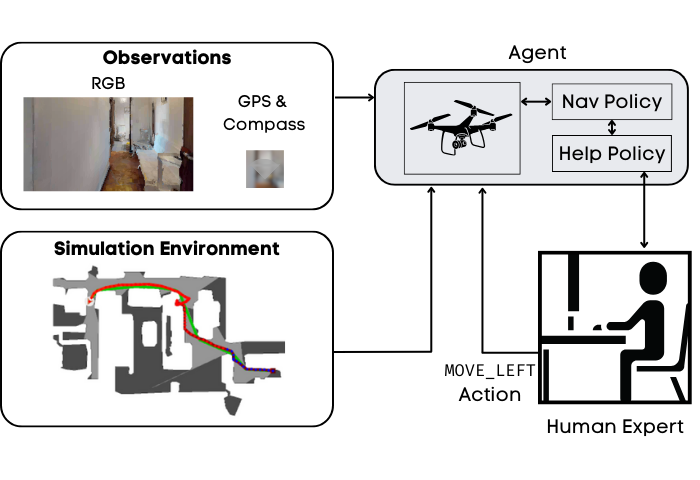}
	\caption{Semi-autonomous system design. The simulation environment generates observations that are given to the autonomous agent, which communicates with a help policy that interacts with the human expert to return a set of optimal actions. The agent then directly interacts with the environment using its own predictions and manual commands from part-time human interaction.}
	\label{fig:approach}
\end{figure}

\section{Approach}
Beginning with a pretrained point navigation policy with frozen weights, we train a secondary policy that learns when to engage a human operator for manual assistance at optimal points in time.
During training, a human expert is unavailable, so we simulate the human expert using a shortest path planner.
The secondary policy uses the navigation policy's encoder output along with goal information as observations and is optimized by rewarding success and penalizing unnecessary human action.
Figure \ref{fig:approach} summarizes this system design.

\subsection{Navigation Policy Design and Optimization}
\subsubsection{Autonomous Policy}
For the base autonomous policy within our semi-autonomous system, we utilize a pretrained actor-critic style point navigation policy provided by Habitat \cite{habitat19iccv} baselines.
This policy takes in \texttt{RGB} imagery as visual input and point goal information (in the form of a relative goal vector including the distance away and orientation towards the goal location) and outputs one of four actions: \texttt{turn\_left}, \texttt{turn\_right}, \texttt{move\_forward}, or \texttt{stop}.
The policy utilizes a ResNet18 \cite{he2016deep} embedding backbone for the \texttt{RGB} input, and the embedded visual input is combined with the point goal information as input to the actor-critic layers.
The actor is a gated recurrent unit and the critic is a fully connected layer.
This policy was optimized using proximal policy optimization \cite{schulman2017proximal} using a navigation reward $r_t$ (Equation \ref{eq:habitatreward}) defined at each time step $t$ as the sum of three terms: a success reward $s$ (which is zero until the goal is reached), the change in distance to goal at time step $t$ ($d_{t-1} - d_{t}$), and a time penalty $\lambda$.
See the Habitat paper for full details.
This policy was pretrained on the Matterport3D (MP3D) dataset \cite{Matterport3D}, which is roughly one order of magnitude smaller than the Habitat-Matterport 3D (HM3D) dataset \cite{ramakrishnan2021habitat} on which we perform our evaluation.
The policy achieves a baseline success rate of 52.3\% on the HM3D dataset. 
\begin{eqfloat}
	\begin{equation}
		r_t = \left\{\begin{matrix}
			s + d_{t-1} - d_t + \lambda & \textrm{if\ goal\ is\ reached} \\
			d_{t-1} - d_t + \lambda & \textrm{otherwise} \\
		\end{matrix}\right. 
		\label{eq:habitatreward}
	\end{equation}
	\caption{Reward function for the baseline autonomous policy, as defined by Habitat \cite{habitat19iccv}. $s$ is a success reward, $d_t$ is the distance to the goal at time $t$, and $\lambda$ is a time penalty that encourages efficiency.}
\end{eqfloat}
\subsubsection{Interaction Policy}
We aim to find an optimal policy that, using the embedded visual input from the fully autonomous policy and point goal information, determines whether or not to engage a human for assistance at time step $t$.
Our objective is to maximize success in task completion while minimizing the ratio of the number of human actions $h_t$ to total actions (the number of human actions $h_t$ plus the number of agent actions $a_t$).
Because visual inputs are already embedded, we use a lightweight multi-layer perceptron for the interaction policy design.

Given visual embedding and point goal observations $\theta_t$, this policy outputs a probability distribution over the actions of the robot proceeding autonomously (action $a_r$) or asking for help (action $a_h$).
We define certainty $c_t$ as the margin between these probabilities, or $c_t = \left| \mathbb{P}(a_h | \theta_t) - \mathbb{P}(a_r | \theta_t) \right|$, with uncertainty defined as $(1-c_t)$.
With this definition, uncertainty in feature extraction becomes an indicator that human interaction will benefit overall success.
This estimate is utilized in our interaction reward function design.
The interaction policy is also optimized using proximal policy optimization, and because this method does not explicitly support multiple rewards, we define a modified reward function to optimize success and interaction efficiency jointly.

Like the navigation policy reward, the interaction policy reward $r^*_t$ (Equation \ref{eq:interactionreward}) is defined as the sum of three terms: the success reward $s$, a cumulative distance reward $d^*_t$, and an interaction penalty $\lambda^*_t$.
$d^*_t$ (Equation \ref{eq:dstar}) is a cumulative distance reward with hyperbolic discounting applied.
We set the discounting factor $\lambda_d$ to be 0.9 during training.
We find optimal performance by explicitly discounting the distance reward in this manner.
$\lambda^*_t$ (Equation \ref{eq:lambdastar}), which is only applied during times of human interaction, is the product of three terms: a hyperparameter $\lambda_h$ weighing the importance of human interaction, our estimate of uncertainty $(1-c_t)$, and the total ratio of human actions to total actions.
\begin{eqfloat}
	\begin{equation}
		r^*_t = \left\{\begin{matrix}
			s + d^*_t - \lambda^*_t & \textrm{if\ goal\ is\ reached} \\
			d^*_t - \lambda^*_t & \textrm{otherwise} \\
		\end{matrix}\right.
		\label{eq:interactionreward}
	\end{equation}
	\caption{Reward function for the interaction policy. $s$ is a success reward. $d^*_t$ is a cumulative distance reward defined in Equation \ref{eq:dstar}. $\lambda^*_t$ is a human interaction penalty defined in Equation \ref{eq:lambdastar}.}
\end{eqfloat}

\begin{eqfloat}
	\begin{equation}
		d^*_t = \sum_{i=0}^{t} \frac{d_t - d_{t-1}}{1 + i\lambda_d}
		\label{eq:dstar}
	\end{equation}
	\caption{Cumulative distance reward applying hyperbolic discounting with the discounting factor $\lambda_d$, which is set to 0.9 during training.}	
\end{eqfloat}

\begin{eqfloat}
	\begin{equation}
		\lambda^*_t = \left\{\begin{matrix}
			\lambda_h \cdot (1-c_t) \cdot \frac{h_t}{h_t + a_t} & \textrm{if\ human\ is\ requested} \\
			0 & \textrm{otherwise} \\
		\end{matrix}\right.
		\label{eq:lambdastar}
	\end{equation}
	\caption{Interaction penalty. If human interaction is requested, the penalty is the product of the importance of human interaction $\lambda_h$, our uncertainty measurement $(1-c_t)$, and the ratio of human actions $h_t$ to total actions $h_t + a_t$. If no human interaction is requested, there is no penalty.}
\end{eqfloat}

For faster optimization we simulate the human expert using a shortest-path planner.
Though humans are not perfect operators, the shortest path is often a reasonable estimate for the path that a human operator would ideally take.
Our interaction policies are optimized for 250,000 time steps, yielding around 1500 episodes in total.

\subsection{Simulation Environments}
We use the AI Habitat simulation environment \cite{habitat19iccv}.
Habitat is a simulation platform for embodied artificial intelligence, including datasets for a variety of navigation tasks grounded in vision and natural language.
The navigation task of interest is point navigation \cite{habitat2020sim2real}, which involves providing an exact 2D location in the environment for a robot to navigate towards given GPS, compass and visual sensor information.
The main challenge of this task is the agent's ability to both navigate around obstacles and determine the shortest path to the goal, avoiding dead ends throughout the path.
We optimize the interaction policy on the MP3D dataset and perform evaluation on the HM3D dataset, which contains a completely unique scene dataset as compared to MP3D.
To further evaluate transfer to unseen environments, we use the FlightGoggles simulation environment \cite{guerra2019flightgoggles}.
The FlightGoggles simulation environment is a photorealistic virtual environment of the Stata Center building at the Massachusetts Institute of Technology.
This virtual environment is nearly identical to the realistic environment given similar lighting conditions, and though it is an indoor commercial space, it is visually unique in comparison to both Matterport derived datasets, providing a unique opportunity to evaluate zero-shot domain adaptation as well as sim-to-real transfer.

\subsection{Human Interaction Design}
\begin{figure}
	\centering
	\includegraphics[width=0.5\textwidth]{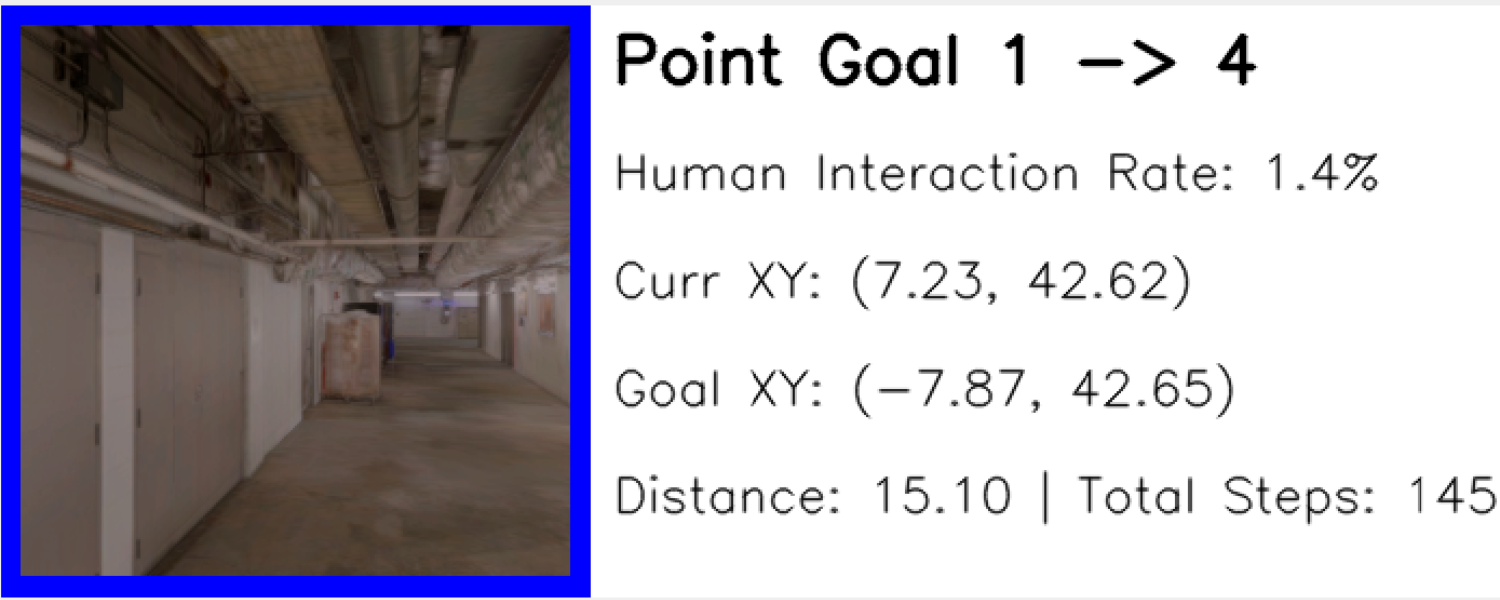}
	\caption{Human interaction interface design. Human experts are presented with the same information as autonomous agents, but presented in a human readable format. Humans are not given top-level maps during navigation.}
	\label{fig:interaction}
\end{figure}

For our evaluation within the FlightGoggles environment, a shortest path planner is unavailable, so we incorporate a real human expert.
This human expert is familiar with the environment, but is unaware of the start and goal positions of the dataset before interaction.
During interaction, the human expert is presented with a graphical user interface providing similar information available to autonomous agents, but in a human-readable format.
Figure \ref{fig:interaction} shows an example of this interface.

The interface shows the current and goal position of the user within the environment, along with progress information including distance to goal, human interaction rates, and total episode steps.
When the agent proceeds autonomously, the image frame has a red border, while during interaction the border is blue, signifying the need for human interaction.
Humans provide manual input using a joystick controller.

\section{Results}
We conduct experiments comparing the overall success and path-length weighted success for autonomous and semi-autonomous control policies within the system.
Our baseline navigation policy is pretrained on the MP3D dataset using \texttt{RGB} image input.
We evaluate control policies using the validation split of the HM3D dataset.
Our evaluation metrics are:
\begin{enumerate}
	\item Success Weighted by Path Length (SPL): scales the success indicator by the actual path's deviation from the optimal path.
	\item Success: achieving a distance of less than 0.25 meters away from the goal point when executing the "stop" action
	\item Human contribution: the ratio of the path affected by human interaction.
	\item Efficiency ($\xi$): the efficiency of semi-autonomous execution, defined as the ratio of increase in navigation success compared to the baseline autonomous policy divided by the rate of human contribution. 
\end{enumerate}

As a baseline, the fully autonomous agent trained on the MP3D dataset achieves success navigating to the goal destination on the HM3D validation split at a rate of 0.523, with an average SPL of 0.426.

\subsection{Semi-Autonomous Navigation}
We first evaluate the performance of the semi-autonomous system.
We compare interaction policies with the human penalty weight ranging from 0 (no penalty) to 1.0 (penalize fully). Table \ref{tab:human} summarizes these results.
The human interaction penalty effectively scales how much human interaction we allow at maximum.
When zero penalty is applied, our policy learns to rely on human interaction while fully maximizing success, while the maximum penalty results in diminishing returns from the efficiency perspective.
\begin{table}
	\centering
	\caption{Examining the impact of human contribution on the overall performance of the semi-autonomous system. We evaluate interaction policies with varying weights for the interaction penalty ($\lambda_h$). When $\lambda_h$ equals zero, no penalty is applied. While penalizing interaction reduces overall success, even small amounts of interaction result in significant improvements in success compared to the autonomous baseline.}
	\begin{tabular}{ccccc}
		\toprule[1pt]
		$\lambda_h$ & SPL $\uparrow$ & Success $\uparrow$ & Interaction Rate $\downarrow$ & $\xi$ $\uparrow$ \\ \dmidrule
		0.0 & \textbf{0.944} & \textbf{0.959} & 0.714 & 0.611\\
		0.5 & 0.786 & 0.906 & 0.300 & \textbf{1.277}  \\
		1.0 & 0.510 & 0.622 & \textbf{0.126} & 0.786 \\ \midrule
		\textbf{Baseline Agent} & 0.426 & 0.523 & - & - \\
		\bottomrule[0.75pt]
	\end{tabular}
	\label{tab:human}
\end{table}
Our most optimal policy results in the highest efficiency rating $\xi$, applying a moderate penalty (0.5) to human interaction resulting in a 0.38 increase in success and a 0.36 increase in SPL with only a 0.30 human interaction rate.
For the optimal policy where $\xi \geq 1$, we observe a roughly linear relationship between human interaction and success improvement, suggesting that the percentage of time spent intervening occurs at points where the agent will directly benefit.

The most significant factor in agent failure stems from choosing sub-optimal actions in favor of exploration and obstacles impacting forward motion.
Because agent actions may result in collisions that prevent movement, traversing some locations within complex scenes can result in agents becoming stuck between obstacles.
Rather than struggling to avoid these tight obstacles, incorporating human intervention allows for agents to be freed quickly.

\begin{figure}
	\begin{subfigure}
		\centering
		\includegraphics[width=0.24\textwidth]{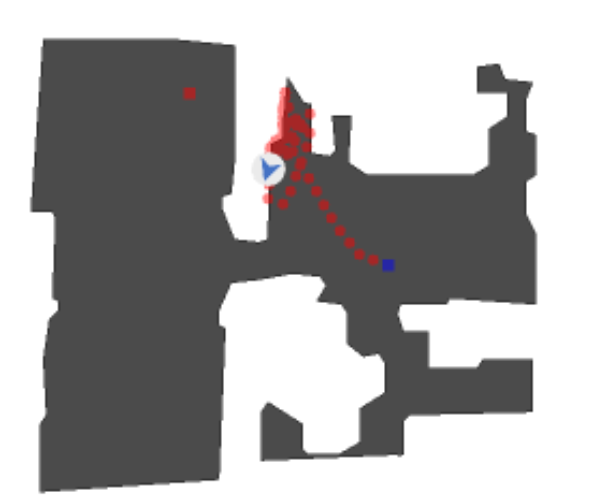}
		\includegraphics[width=0.24\textwidth]{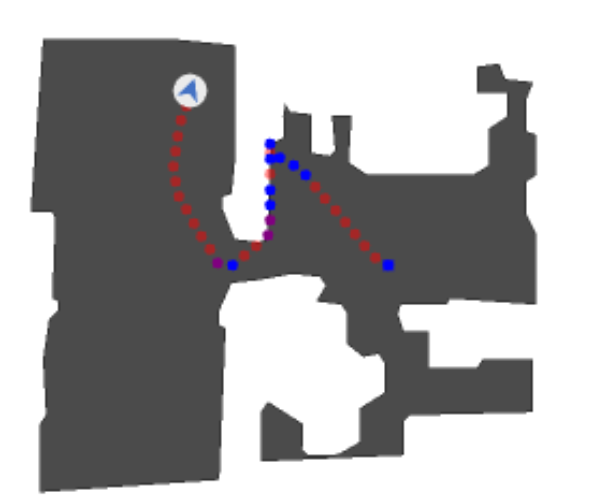}
		\caption*{(a) Without intervention, the robot attempts to take the shortest path to the goal, looping back and forth and getting stuck at a dead end.}
	\end{subfigure}
	\begin{subfigure}
		\centering		\includegraphics[width=0.24\textwidth]{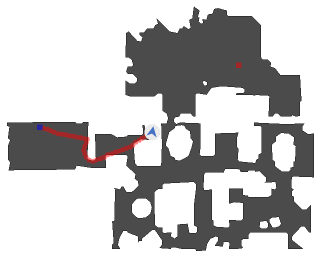}
		\includegraphics[width=0.24\textwidth]{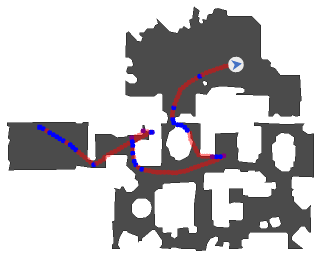}
		\caption*{(b) Without intervention, the robot gets stuck at a tight obstacle.}
	\end{subfigure}
	\caption{Examples of the impact of uncertainty estimation in the selection of optimal points of expert intervention. Pairs of interactions are side by side, with autonomous execution on the right. Rather than getting stuck within dead ends, loops, or tight sets of obstacles, expert intervention avoids these areas of catastrophic failure.}
	\label{fig:fig6}
\end{figure}

\begin{figure}
	\begin{subfigure}
		\centering
		\includegraphics[width=0.22\textwidth]{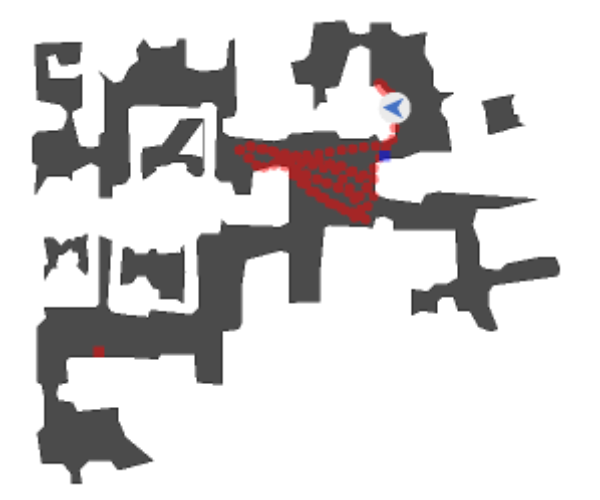}
		\includegraphics[width=0.22\textwidth]{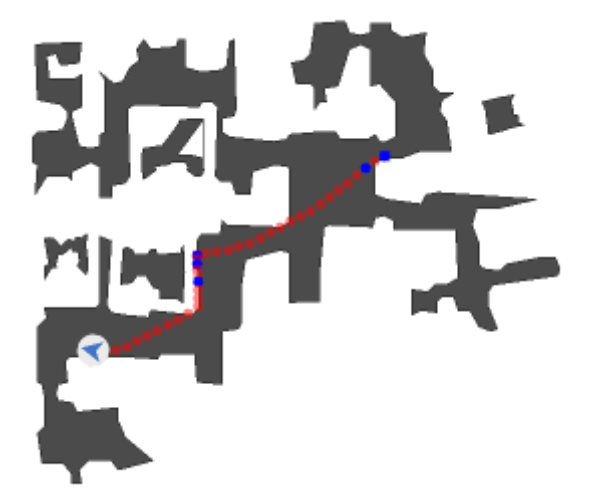}
		\caption*{(a) Because a direct path to the goal is obstructed, the robot wastefully explores path in the wrong direction without intervention.}
	\end{subfigure}
	\begin{subfigure}
		\centering
		\includegraphics[width=0.22\textwidth]{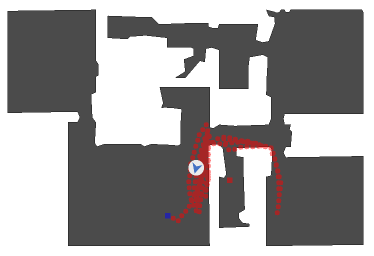}
		\includegraphics[width=0.22\textwidth]{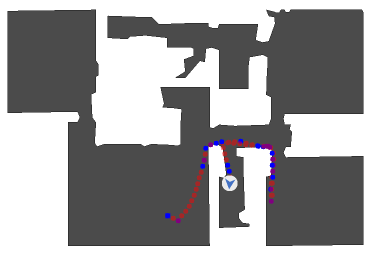}
		\caption*{(b) The robot attempts to explore the larger rooms autonomously, it does not find the smaller room containing the goal without human intervention.}
	\end{subfigure}
	\caption{Human intervention reducing wasteful exploratory behavior. When agents struggle to progress towards the goal location, exploration is sometimes effective, but may become wasteful in more complex scenes. Relying on expert interaction significantly reduces this waste.}
	\label{fig:fig5}
\end{figure}

Figure \ref{fig:fig6} shows examples of expert interaction releasing an agent from becoming stuck in multiple areas that result in failure during the autonomous case.
Our uncertainty-based interactive approach avoids these catastrophic failure cases.
Further, interaction allows agents to avoid wasteful exploration. Figure \ref{fig:fig5} shows the agent engaging in autonomous exploration, a common learned behavior when progress towards the goal location is not achieved by proceeding down a clearly unobstructed path.
When an agent is uncertain, this exploratory behavior can be exceedingly wasteful.
In the semi-autonomous case, we see that with just a few points of expert intervention, we are able to avoid this exploratory behavior and proceed towards the goal in an optimal fashion.

\begin{figure}
	\centering
	\begin{subfigure}
		\centering
		\includegraphics[width=0.24\textwidth]{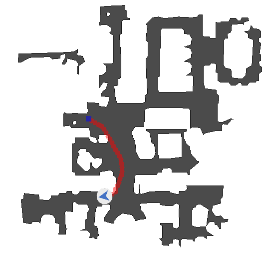}
		\includegraphics[width=0.24\textwidth]{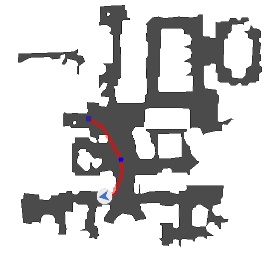}
		\caption*{(a) With a largely unobstructed path, the robot has high certainty and requests minimal intervention.}
	\end{subfigure}
	\begin{subfigure}
		\centering
		\includegraphics[width=0.24\textwidth]{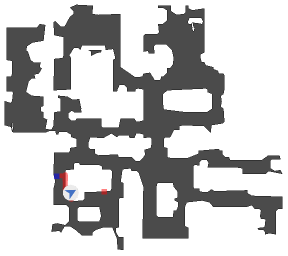}
		\includegraphics[width=0.24\textwidth]{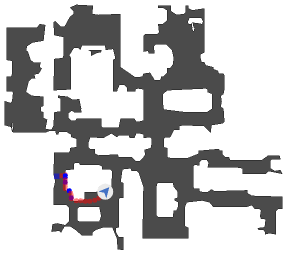}
		\caption*{(b) With a very short path, uncertainty is minimal (except at areas of tight obstruction) and minimal intervention is requested.}
	\end{subfigure}
	\caption{Our uncertainty based policy avoids requesting human intervention when intervention is unnecessary for success.}
	\label{fig:fig4}
\end{figure}

We also note that our interaction policy refrains from requesting human assistance during times of certainty.
Figure \ref{fig:fig4} shows examples of this optimal semi-autonomous control.
Though human assistance is requested at a few number of points during execution, because of high confidence and consistent progress towards the goal location, intervention is largely avoided.

\subsection{Zero-Shot Transfer with a Real Human Expert}
Finally, we deploy this semi-autonomous system with a real human expert within the FlightGoggles simulation environment showing transfer with zero further refinement of any policy.
We create a small test dataset of 14 episodes navigating within the basement environment.
We evaluate overall success in navigating within the environment for both autonomous and semi-autonomous interaction.
Table \ref{tab:fg} summarizes these results.
\begin{table}
	\centering
	\caption{Zero-shot transfer performance in the Stata Center Environment. Although the baseline policy does not transfer to this environment, interaction at a similar rate during training allows for optimal performance in the semi-autonomous context.}
	\begin{tabular}{cccc}
		\toprule[1pt]
		Policy & Success $\uparrow$ & Interaction Rate $\downarrow$ & $\xi$ $\uparrow$ \\ \dmidrule
		Semi-Autonomous & 0.929 & 0.232 & 3.92 \\ \midrule
		\textbf{Autonomous: 14 episodes} & 0.07 & - & - \\	
		\textbf{Autonomous: 100 episodes} & 0.09 & - & - \\
		\bottomrule[0.75pt]
	\end{tabular}
	\label{tab:fg}
\end{table}
We first note that the autonomous baseline does not transfer to this environment.
On the test set, the autonomous baseline achieves success navigating to the goal location at a rate of only 0.07, and when extending this dataset to 100 unique episodes, that rate increases to only 0.09.
However, when adding a real human expert, we are able to achieve a success rate of 0.92 with an interaction rate of only 0.23, showing the power of human intervention to help poorly performing policies achieve optimal performance.
\section{Future Work}
Though this simple approach to part-time interaction showed improved success for autonomous agent navigation, we have yet to explore a number of approaches that provide improved support from human experts while reducing the interaction burden.
First, as our interaction approach requires manual control from expert operators, human operators need to pay attention to task execution to provide optimal support for autonomous agents.
As we think about extending this work to more complex tasks like object search and rescue as well as multi-agent systems, more optimal approaches for interaction and intervention are of interest.
In particular, the ability for agents to ask questions and receive answers in natural language that can condition autonomous policies along with the ability for operators to provide high level verbal commands such as \texttt{check in that room} will be exceedingly more efficient than manual control hand-off.
Furthermore, human experiments are necessary to evaluate optimal interaction approaches and inform the design of efficient interfaces.

The refinement of our existing navigation policies is also of great interest.
As our interactive policies identify areas of confusion within the policy, there is a large opportunity to minimize areas of failure within our baseline policy in order to improve autonomous performance.
As human interaction is part-time and limited, this may be a difficult challenge.
Because the optimization of large scale navigation policies typically requires large simulation environments or large demonstration datasets, the number of part-time demonstrations that would be necessary to optimally refine a pretrained navigation policy for inference in a new environment is likely infeasible to capture during a small number of episodes, and would likely result in over-fitting to a few episodes within the target domain.
Therefore, the exploration of few-shot optimization and meta refinement approaches are of interest and will be explored in future work.
 
\section{Conclusion}
In this work, we present a semi-autonomous system for interactive point-goal navigation.
Motivated by the unreliability of fully autonomous systems operating in unfamiliar domains, we design a system that improves real-time performance through lightweight, part-time human interaction.
We develop an external interaction policy that re-purposes visual navigation features to optimally learn when to ask for help in unfamiliar environments.
Our interaction policies incorporate implicit measures of uncertainty that determine when human interaction will directly improve performance, enabling significant improvement in task completion with minimal human interaction.

Our best performing policy using this approach yields a 0.9 success rate on our validation set, with a 0.38 increase in success compared to the baseline (0.906 vs. 0.523) with an intervention rate of 0.3.
When transferring this policy using a zero-shot approach to a completely unfamiliar environment, we are able to improve success from less than 0.1 to 0.92 with a real human expert.
This approach shows promise for implementing lightweight policies that leverage part-time human interaction for improved autonomous performance, without an explicit understanding of state or task information.

\section{Acknowledgment}
Research was sponsored by the United States Air Force Research Laboratory and the Department of the Air Force Artificial Intelligence Accelerator and was accomplished under Cooperative Agreement Number FA8750-19-2-1000. The views and conclusions contained in this document are those of the authors and should not be interpreted as representing the official policies, either expressed or implied, of the Department of the Air Force or the U.S. Government. The U.S. Government is authorized to reproduce and distribute reprints for Government purposes notwithstanding any copyright notation herein.

\bibliographystyle{IEEEtran}
\bibliography{./references}

\end{document}